\documentclass[A4paper, 11 pt, conference]{ieeeconf}  
\IEEEoverridecommandlockouts       
\usepackage{geometry}
\usepackage{graphicx}
\usepackage{amsfonts}
\usepackage{amsmath}
\usepackage{amssymb}
\usepackage{color}
\usepackage{cite}
\usepackage{graphicx,color,overpic}
\usepackage{psfrag}
\usepackage{amssymb}
\usepackage{times}
\usepackage{latexsym}
\usepackage{bm}
\usepackage{cases}
\usepackage{array}
\usepackage{fancyhdr}
\usepackage{setspace}
\usepackage{subfigure}
\usepackage{url}
\usepackage{multirow}
\usepackage{epstopdf}
\usepackage[ruled,linesnumbered]{algorithm2e}
\usepackage{epsfig}
\usepackage{fancybox}
\usepackage{textcomp}
\usepackage{algpseudocode}

\DeclareMathOperator*{\argminA}{arg\,min} 
\newcommand{\x}[0]{\mathbf{x}}

\newcommand{\f}[0]{\mathbf{f}}
\newcommand{\e}[0]{\mathbf{e}}
\newcommand{\C}[0]{\mathbf{C}}
\newcommand{\SIM}[0]{\mathrm{sim}}
\newcommand{\binning}[0]{\mathbf{b}}

\begin{document}

\title{\LARGE \bf
Collaborative Radio SLAM for Multiple Robots based on WiFi Fingerprint Similarity}

\author{Ran Liu, Zhenghong Qin, Hua Zhang, Billy Pik Lik Lau, Khairuldanial Ismail, \\
Achala Athukorala, Chau Yuen, Yong Liang Guan, and U-Xuan Tan
\thanks{R. Liu, B. P. L. Lau, K. Ismail, A. Athukorala, C. Yuen, and U-X. Tan are with the Engineering Product Development Pillar, Singapore University of Technology and Design, 8 Somapah Rd, Singapore, 487372. {\{\tt\small ran\_liu\}@sutd.edu.sg}.
}
\thanks{R. Liu, Z. Qin, and H. Zhang are with the School of Information Engineering, Southwest University of Science and Technology, Mianyang, China, 621010.
}
\thanks{Y. L. Guan is with the School of Electrical and Electronic Engineering, Nanyang Technological University, 50 Nanyang Avenue, Singapore, 639798.
}
\thanks{This work is supported by the National Key R\&D Program of China (2019YFB1310805) and the National Science Foundation of China (12175187).}
}
\maketitle 
\thispagestyle{empty}

\begin{abstract}
Simultaneous Localization and Mapping (SLAM) enables autonomous robots to navigate and execute their tasks through unknown environments. 
However, performing SLAM in large environments with a single robot is not efficient, 
and visual or LiDAR-based SLAM requires feature extraction and matching algorithms, which are computationally expensive.
In this paper, we present a collaborative SLAM approach with multiple robots using the pervasive WiFi radio signals. 
A centralized solution is proposed to optimize the trajectory based on the odometry and radio fingerprints collected from multiple robots. 
To improve the localization accuracy, a novel similarity model is introduced that combines received signal strength (RSS) and detection likelihood of an access point (AP). 
We perform extensive experiments to demonstrate the effectiveness of the proposed similarity model and collaborative SLAM framework.
\end{abstract}

\section{Introduction}
\label{sec:introduction}
The advancement of Simultaneous Localization and Mapping (SLAM) is the key enabler for autonomous navigation, as it deals with self-localization and building a map of unknown environment \cite{Yassin2017Recent,Lau2019survey}.
Many solutions have been proposed to address this problem in the robotics community during the past decades. 
With the emergence of low-cost sensors and increasing process power of modern devices, 
one is able to perform SLAM online and several commercial products (for example Intel RealSense and Google ARCore) are available in the market.


It has been demonstrated that in a single robot application, SLAM is capable of achieving substantial robustness and accuracy \cite{MurArtal2015ORB, ran_mechatronics}.
With such achievement, the robotics community research focus has been shifting from high accuracy to high efficiency, resulting usage of multi-robot to perform SLAM.
The advantages of leveraging such a technique allow a more robust system to be created by allowing sensing information to be shared among the cluster of robots. 
It boosts the overall efficiency of a mission by further dividing the SLAM tasks among robots, where a single robot might consume a longer duration to optimize the trajectory. 
That being said, the multi-robot approach has a few challenges that need to be addressed, such as network reliability, consistency, and information sharing policy.

\begin{figure}
\centering
\includegraphics[width=0.45\textwidth]{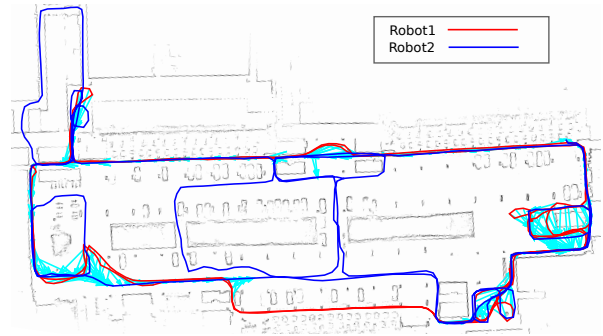}
\caption{
Overview of our collaborative radio SLAM with two robots. We perform SLAM by fusing odometry and radio fingerprints collected from multiple robots. Fingerprint-based constraints between the two robots are shown in cyan.
}
\label{fig_radio_lidar_slam}
\vspace{-0.7cm}
\end{figure}
Collaborative SLAM is one of the common methods leveraging the collaborative nature of multiple robots to perform SLAM.
Recent works have been focusing on using different robots for tracking and mapping in unknown environments using unmanned aerial vehicle \cite{Zhang2018Cloud}, mobile devices \cite{Liu2020Cooperative,Liu2020Collaborative}, and autonomous robots \cite{ran_iros2021}.
Depending on the types of sensors, the complexity of the computation can be high and therefore not suitable for real-time sensing operation.
Therefore, selecting a set of useful sensors is essential for developing fast and efficient SLAM.

Nowadays, WiFi networks are abundant in a dense urban area, which are commonly used for data transmission or performing localization in the area where GPS signal is not accessible. 
It offers low hardware requirement and computational cost due to its ubiquitous in-built sensing capabilities as reviewed in \cite{Yassin2017Recent}.
Hence, analytical models that represent the radio signal distribution are commonly used for trajectory optimization through SLAM \cite{Ferris2007Wifi, Huang2011Efficient}. 
However, they suffered from multipath issues in an uncontrolled environment and hence not suitable in the real-world. 
That being said, it can be further improved using radio fingerprint \cite{Zhou2018Robust} resulting in less signal distortion when performing localization. 
Despite being a less radio distortion approach, a single robot may take a longer time to optimize the pose estimation, which might not be suited for large scale environments.
Thus, it motivates us to propose the collaborative radio SLAM in a multi-robot scenario.

In this paper, we present collaborative radio SLAM that performs simultaneous localization and mapping based on the pervasive WiFi signals in a multi-robot scenario, as shown in Figure \ref{fig_radio_lidar_slam}. We investigate the collaborative nature of multiple robots when they actively listen to the surrounding WiFi received signal strength (RSS) measurements from the existing WiFi access points (APs) in an infrastructure. 
To improve the positioning accuracy, we present a novel similarity measure that considers the received signal strength and the detection likelihood of an AP. 
Note that this method does not require prior map knowledge nor the access points location, which makes it more suitable for indoor buildings with WiFi network deployed.
Using the mutual information concept, 
the proposed collaborative SLAM solution allows us to improve the pose estimation accuracy by incorporating the measurements from multiple robots. 
The contributions of this paper are three folded and are listed as follows:
\begin{itemize}
	\item We propose the collaborative radio SLAM to optimize the trajectory using a multiple-robot scenario based on fingerprint similarity by promoting cooperation among multiple robots. 	
	\item A new similarity measure that combines the received signal strength and the detection likelihood of the access point is proposed to improve the localization accuracy. 
	\item We perform extensive experiments to validate the proposed similarity measure as well as our proposed collaborative SLAM solution.
\end{itemize}

This paper is organized as follows:
Section \ref{sec:approach} presents the collaborative mechanism and its theoretical foundation. 
We then show the experimental results in Section \ref{sec:experiment} for different case studies and analyze them. 
Lastly, we conclude our works in Section \ref{sec:conclusion}.

\section{Collaborative Radio SLAM with Fingerprint Similarity}
\label{sec:approach}
Figure \ref{fig_radio_lidar_slam} shows a scenario of two robots perform SLAM in an indoor environment with WiFi. 
Each robot provides odometry and carries mobile phones to scan the received signal strength of the access point in the environment.
The purpose is to perform SLAM by fusing the odometry and radio measurement among all robots without prior knowledge about the infrastructure. While we show only two robots for illustration, the algorithm and discussion are suitable for the general case of multiple robots.

\subsection{Problem Formulation}
\label{sec:formulation}
Formally, the goal is to estimate the path of robot $k$ up to time $T$: $\x_k^{1:T}=\{\x_k^1,...,\x_k^T\}$, where ${{\x}_k^{(t)}} = {[x_k^{(t)},y_k^{(t)},\theta_k^{(t)}]}$ denotes the pose of the robot $k$ at time $t$.
The radio measurement scanned by robot $k$ at time $t$ is denoted as $\f_k^t$. 
Our goal is to estimate the trajectory of all robots given the odometry and radio measurements collected by all robots. 
We consider a centralized solution, since the data volume of radio measurements is small when compared to the visual images or LiDAR scans.

In general, graph-based SLAM uses the pose of the robot to represent a node. The measurements are decoded as constraints in the graph. 
The graph-based SLAM aims to find the best configuration of the poses to minimize the error of all constraints through maximum likelihood estimation. In the context of our collaborative SLAM, the constraints consist of three sources: 1) the odometry-based constraints, 2) the individual radio fingerprints-based constraints, and 3) the constraints obtained by comparing the similarity of radio fingerprints collected from different robots. The problem is turned into finding the best poses ${\x^*}$ to minimize the following equation:
\small
\begin{equation}
\begin{split}
&\argminA_{\x}  \sum_{k=1}^K \sum_{t=1}^{T} \underbrace{ {\e} ({\x_k^{t-1},\x_k^{t}, \Delta \x_k^{t} })^T \Omega_{k}^{t}  {\e} ({\x_k^{t-1},\x_k^{t}, \Delta \x_k^{t} }) }_{\text{Odometry-based constraints}}+\\
&\sum_{k=1}^K \sum_{(\x_k^i,\x_k^j) \in \C_k} \underbrace{ {\e} ({\x_k^{i},\x_k^{j}, d(\f_k^{i},\f_k^{j}) })^T \Omega_{k}^{i,j}  {\e} ({\x_k^{i},\x_k^{j}, d(\f_k^{i},\f_k^{j}) }) }_{\text{Individual radio fingerprint-based constraints}}\\
   &+\sum_{(\x_k^i,\x_l^j) \in \C_{k,l}} \underbrace{ {\e} ({\x_k^{i},\x_l^{j}, d(\f_k^{i},\f_l^{j}) })^T \Omega_{k,l}^{i,j}  {\e} ({\x_k^{i},\x_l^{j}, d(\f_k^{i},\f_l^{j}) }) }_{\text{Radio fingerprint-based constraints between robots}}\\
 \end{split}
\label{eq:optimization}
\end{equation}
\normalsize
where $K$ is the number of robots moving in the environment. $\e(\cdot)$ denotes the error function, which is computed based on the given poses and the constraints inferred from the observations (i.e., odometry and radio fingerprints). The former is represented as the sequential odometry measurements (a 3$\times1$ vector including the 2D displacement and the rotation between $\x_k^{t-1}$ and $\x_k^{t}$).  
The latter is represented as the fingerprint-based constraint (i.e., a distance measurement $d(\f_k^{i},\f_l^{j})$ that is determined by the similarity of two radio fingerprints $\f_k^{i}$ and  $\f_l^{j}$, see Section \ref{sec:model_training}). 
Similar to our previous work in \cite{Liu2020Collaborative}, we train a model to feature the relationship between the distance and the fingerprint similarity.
To obtain a better model, we propose a novel similarity measure that additionally considers the detection likelihood of an AP (see Section \ref{sec:radio_similarity} for details). 
Constraints are additionally parameterized with a certain degree of uncertainty, which is denoted as the information matrix (i.e., $\Omega_{k}^{t}$, $\Omega_{k}^{i,j}$, and $\Omega_{k,l}^{i,j}$ ) in Equation \ref{eq:optimization}. 
$\C_k$ denotes the set containing the pairs of fingerprint constraints in individual robot $k$ 
and $\C_{k,l}$ represents the set of fingerprint constraints between robot $k$ and robot $l$. 
\small
\begin{equation}
 \begin{split}
&s_r=\prod_{n=1}^{H} \exp(-\frac{(f_{k,n}^i-f_{l,n}^j)^2}{2\sigma_r^2}),\\   
 \end{split}
\label{eq:rss}
\end{equation}
\normalsize
where $\sigma_r^2$ represents the received signal variance, which is set to 36dBm according to \cite{Berkvens2014BioWiFiSLAM}. 
The detection likelihood of extra APs is modeled as:
\small
\begin{equation}
 \begin{split}
&s_\tau=\prod_{n=1}^{M_k^i} \exp(-\frac{(\tau_{k,n}^{i})^2}{2\sigma_\tau^2}) \prod_{n=1}^{M_l^j} \exp(-\frac{(\tau_{l,n}^{j})^2}{2\sigma_\tau^2}),\\   
 \end{split}
\label{eq:detection}
\end{equation}
\normalsize
where $\sigma_\tau$ is parameter to control the weights of detection likelihood. 
The similarity of two fingerprints is modeled as a combination of RSS likelihood of common APs and detection likelihood of extra APs:
\small
\begin{equation}
 \begin{split}
&s=\SIM(\f_{i}^m,\f_{j}^n)=s_\tau \cdot {\sqrt[H]{s_r}}  \cdot \frac{H}{H+M_k^i+M_l^j}\\   
 \end{split}
\label{eq:similarity}
\end{equation}
\normalsize
Removing the first term (i.e., $s_\tau$) in Equation \ref{eq:similarity} leads to the Gaussian similarity measure. 
Our novel similarity measure penalizes the APs with low detection probability and gives better localization results as shown in the experimental results in Section \ref{sec:experiment}.
\subsection{Fingerprint Similarity with RSS and Detection Likelihood}
\label{sec:radio_similarity}
Radio fingerprinting represents location with radio signals from surrounding WiFi APs, 
which are shown to be robust against location-dependent distortions.
The radio fingerprint $\f_k^t$ recorded at $\x_k^t$ consists of the average RSS values $\{ f_{k,1}^t,...,f_{k,N}^t \}$ and the detection probability $\{ \tau_{k,1}^t,...,\tau_{k,N}^t \}$ in a given scanning period, i.e., 5 seconds in our case. 
Given two fingerprints $\f_k^{i}$ and $\f_l^{j}$, we use $H$ to represent the number of common APs (i.e., the APs appear in both fingerprints). 
In addition, let $M_k^i$ and $M_l^j$ denote number of extra APs (i.e., the APs appear in only one fingerprint) in $\f_k^i$ and $\f_l^j$, respectively.
The similarity function $\SIM(\f_{k}^i,\f_{l}^j)$ yields a positive value between 0 and 1, 
that considers the signal strength likelihood and the detection likelihood:
\subsection{Model Training}
\label{sec:model_training}
To optimize Equation \ref{eq:optimization}, 
we need to know the distance and the variance of two locations based on radio fingerprints. 
Following our previous work in \cite{Liu2020Collaborative}, we train a model to represent the distance and the variance given a similarity by passing over the measurement collected by the robot (i.e., odometry and radio fingerprints), based on the assumption that the odometry is accurate at a short distance (for example 100m).

We first compute similarity for close fingerprint pairs according to Equation \ref{eq:similarity}. 
These values are annotated with the distance between the two locations obtained by the odometry. 
As a result, we obtain a set of $W$ training samples: $\{s_w,d_w\}_{w=1}^{W}$, 
where $s_w$ and $d_w$ denote the similarity and the distance of a fingerprint pair, which are measured by a similarity measure and the odometry, respectively.
We then train a model which characterizes the distance $\hat{d}(s)$ and the variance $\hat{\sigma}^2(s)$ given a similarity $s$ by binning:
\small
\begin{equation}
 \begin{split}
&\hat{d}(s)=\frac{1}{ c (\binning (s,r)) } \sum_{s-\frac{r}{2} \le s_w \le s+\frac{r}{2}} { d_w}\\    
   &\hat{\sigma}^2(s)=\frac{1}{ c (\binning (s,r)) } \sum_{s-\frac{r}{2} \le s_w \le s+\frac{r}{2}} { (d_w-\hat{d}(s))}^2
\\
 \end{split}
\label{eq:modeling}
\end{equation}
\normalsize
where $c(\binning(s,r))$ counts the number of samples in interval $r$ given a similarity value $s$.
$\hat{d}(s)$ and $\hat{\sigma}^2(s)$ represent the expected distance and variance given a similarity $s$.
\begin{algorithm}
\small
\label{particle_filter_algo}
\SetKw{KwWith}{with}
\SetKw{KwEach}{each}
\KwData{Odometry of $K$ robots $\{\x_1^{1:T}, ..., \x_K^{1:T}\}$ and the associated radio fingerprints $\{\f_1^{1:T}, ..., \f_K^{1:T}\}$}
\KwResult{Optimized trajectory $\x^*$}
\caption{Proposed collaborative SLAM}
$\mathcal{C} \leftarrow \varnothing$ \tcp{Initial constraint set}
\tcp{Odometry-based constraints}
\For{$k\leftarrow1$ \KwTo $K$}
{
\For{$t\leftarrow1$ \KwTo $T$}
{
$\mathcal{C} \leftarrow \mathcal{C} \cup \langle {\e} ({\x_k^{t-1},\x_k^{t}, \Delta \x_k^{t} }),\Omega_{k}^{t} \rangle$
}
}
\tcp{Fingerprint constraints for individual robot}
\For{$k\leftarrow1$ \KwTo $K$}
{
{
\For{$i\leftarrow1$, $j\leftarrow1$ \KwTo $T$ \KwWith $j < i$}
{
\tcp{Similarity of $\f_{k}^i$  and $\f_{k}^j$}
$s \leftarrow s_\tau \cdot {\sqrt[H]{s_r}}  \cdot \frac{H}{H+M_k^i+M_k^j}$ (Equation \ref{eq:similarity}) \\
\If{$\hat{d}(s) < \nu_s$}{$\mathcal{C} \leftarrow \mathcal{C} \cup \langle {\e} ({\x_k^{i},\x_k^{j}, \hat{d}(s) }),\frac{1}{\hat{\sigma}(s)^2}  \rangle$}

}
}
}
\tcp{Fingerprint constraints between different robots}
{
\For{$k\leftarrow1$, $l\leftarrow1$ \KwTo $K$ \KwWith $l \ne k$}
{
\For{$i\leftarrow1, j\leftarrow1$ \KwTo $T$}
{
\tcp{Similarity of $\f_{k}^i$ and $\f_{l}^j$}
$s \leftarrow s_\tau \cdot {\sqrt[H]{s_r}}  \cdot \frac{H}{H+M_k^i+M_l^j}$ (Equation \ref{eq:similarity}) \\
\If{$\hat{d}(s) < \nu_p$ }
{$\mathcal{C} \leftarrow \mathcal{C} \cup \langle {\e} ({\x_k^{i},\x_l^{j}, \hat{d}(s) }),\frac{1}{\hat{\sigma}(s)^2}  \rangle$}

}

}
}
$\rhd$   Pose graph optimization given the constraints $\mathcal{C}$ according to Equation \ref{eq:optimization}
\end{algorithm}
\subsection{Loop Closure Detection}
\label{sec:loop_closure}
To optimize the trajectory with graph-based SLAM, we need to find the fingerprint-based constraints (i.e., loop closures) for individual robot and the constraints between different robots. 
For the individual robot, 
we first check the distance travelled by the robot between two recorded fingerprints. 
If it is larger than a pre-defined threshold (100 meters), we compute the similarity based on Equation \ref{eq:similarity} and obtain the expected distance based on the model in Section \ref{sec:model_training}. 
The reason to exclude the close fingerprints for loop closure detection is that 
the distance measured from odometry in a short period is much precise than the predicted distance obtained from our similarity model in Section \ref{sec:model_training}.
If the expected distance is smaller than a threshold $\nu_s$, we add this constraint to the graph for further optimization. 
Similarly, to detect the fingerprint constraints between different robots, we check the similarity of the fingerprints recorded for different robots. 
If the expected distance is smaller than a threshold $\nu_p$, we add this constraint to the graph. 
The pose graph is optimized based on g2o (General Graphic Optimization) \cite{Grisetti2010Tutorial} with an in-build Levenberg-Marquardt solver. 
An overview of the algorithm is shown in Algorithm \ref{particle_filter_algo}.

\begin{figure}
\centering
\includegraphics[width=0.4\textwidth]{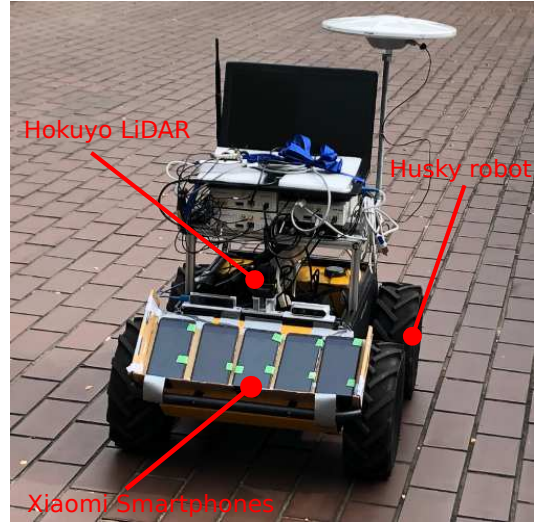}
\caption{
Robot and the sensors used for data collection.}
\label{fig_robot_platform}
\vspace{-0.4cm}
\end{figure}

\section{Experimental Result}
\label{sec:experiment}
In this section, we present the experimental results to validate the proposed approach. 
In particular, Section \ref{sec:experiment_setup} describes the experimental setup and 
Section \ref{sec:experiment_similarity} compares the performance under different similarity measures. 
Section \ref{sec:experiment_clam} presents the results of our proposed collaborative radio SLAM. 
\subsection{Experimental Setup}
\label{sec:experiment_setup}
We evaluated the proposed approach at Nanyang Technological University (NTU) Basement 2 carpark B (approx. 5100 m$^2$).
The experiments were performed using a Clearpath Husky unmanned ground vehicle, as shown in Figure \ref{fig_robot_platform}.
The robot started at different positions and two test sets (namely, Test I and Test II) were recorded. 
The robot moved at an average speed of 0.4m/s. 
Wheel odometry measurements were recorded at a frequency of 10Hz.
Five Xiaomi Mi Max3 smartphones were placed on the robot to scan for WiFi access points at a frequency of 0.5Hz.
We used a time window of 5 seconds to group the scans from the five phones. 
We compute average RSS and count detection probability of an AP to create a fingerprint.
A Hokuyo 2D LiDAR UST-20LX was mounted on the robot to provide the ground truth through Adaptive Monte Carlo Localization (AMCL) based on a map created by GMapping. 

\begin{figure} 
\centering     
        \subfigure[Scatter plot and the model trained using our proposed similarity measure with $\sigma_r=6$ and $\sigma_\tau=4$]{
\label{fig:scatter}
        \includegraphics[width=0.47\textwidth]{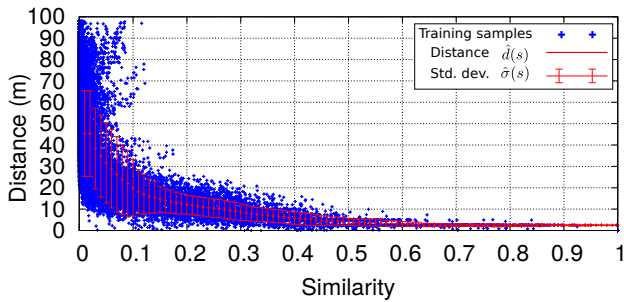}
        }
        \subfigure[Comparison of the similarity models under different similarity measures]{
\label{fig:model_different_sim}
        \includegraphics[width=0.46\textwidth]{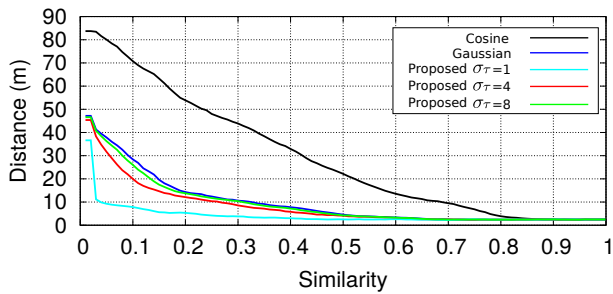}
        }
        \caption[Estimated trajectory]
{Illustration of the similarity models.}
\label{fig_model}
\vspace{-0.4cm}
\end{figure}

We use a binning size $r=0.05$ to generate the model.
The samples used for training and our proposed similarity model is visualized in Figure \ref{fig:scatter}. 
The models generated with different similarity measures are shown in Figure \ref{fig:model_different_sim}.  
As can be seen from this figure, 
the similarity value computed by our proposed approach is smaller than the Gaussian similarity model, which does not consider the detection likelihood, 
as our approach additionally penalizes the detection of extra APs. 
With a large $\sigma_\tau$, the proposed model is equivalent to the Gaussian similarity model. 

\subsection{Evaluation of the Proposed Similarity Model}
\label{sec:experiment_similarity}
In this series of experiments, we compared the localization accuracy between the proposed similarity model and the existing similarity models, namely Gaussian and cosine.
We fix $\sigma_r=6$ for this series of experiments. 
Table \ref{table_models} shows the results of single robot SLAM. 
As can be seen from this table, our proposed similarity model outperforms cosine and Gaussian similarity models in both datasets. 
With a setting of $\sigma_\tau=4$, our proposed model gives the best localization accuracy in both datasets. 
A too large or too small $\sigma_\tau$ leads to decrease in localization accuracy. 
With our proposed model ($\sigma_\tau=4$), we achieve a localization accuracy of 3.261m and 3.364m for Test I and Test II, respectively, 
which is an improvement of 10.8\% and 8.1\% when compared to the Gaussian similarity model (i.e., an accuracy of 3.656m and 3.661m for Test I and Test II). 

In general, too large or too small $\nu_s$ leads to a worse localization result. 
The best setting of $\nu_s$ varies for different datasets. 
For example, $\nu_s=35$ gives the best localization accuracy for Test I, while the best choice for Test II is $\nu_s=20$.
One reason to explain this is that the odometry exhibits different errors for the two datasets, 
which requires a different number of constraints (i.e., obtained by different settings of $\nu_s$) to compensate for the error. 

\begin{table*}[]
\small
\centering
\caption{A comparison of the trajectory estimation accuracy (in meters) under different similarity models with a single robot SLAM.}
\label{table_models}
\begin{tabular}{|c|c|c|c|c|c|c|c|c|c|c|}
\hline
\multirow{2}{*}{Experiment} & \multirow{2}{*}{Approach} & \multicolumn{9}{c|}{Setting of parameter $\nu_s$}                                        \\ \cline{3-11} 
                            &                            & 0      & 5       & 10    & 15    & 20     & 25      & 30      & 35       &  40    \\ \hline
\multirow{5}{*}{Test I}      & Cosine                    & 19.251 & 4.876   & 4.723 & 4.746 & 4.786  & 4.691 & 4.782  & 4.752   & 4.563 \\ \cline{2-11} 
                            & Gaussian                   & 19.251 & 4.573   & 4.375 & 4.234 & 4.062  & 3.943 & 3.656  & 3.849   & 9.346   \\ \cline{2-11} 
                            & Ours with $\sigma_\tau$=1 & 19.251 & 4.754   & 4.337 & 4.371 & 4.370  & 4.370 & 4.370  & 4.370   & 4.543 \\ \cline{2-11} 
                            & Ours with $\sigma_\tau$=4 & 19.251 & 4.503   & 4.195 & 4.167 & 3.995  & 3.853 & 3.618  & \textbf{3.261}   & 3.858    \\ \cline{2-11} 
                            & Ours with $\sigma_\tau$=8 & 19.251 & 4.552   & 4.285 & 4.199 & 4.085  & 3.879 & 3.623  & 3.665   & 5.766    \\ \hline \hline
\multirow{5}{*}{Test II}      & Cosine                   & 10.124 & 9.137   & 4.920 & 4.172 & 3.838  & 3.989 & 6.124  & 8.788   & 12.016     \\ \cline{2-11} 
                            & Gaussian                   & 10.124 & 5.916   & 4.181 & 3.661 & 3.951  & 7.074 & 11.97  & 18.939  & 24.913      \\ \cline{2-11} 
                            & Ours with $\sigma_\tau$=1 & 10.124 & 4.816   & 4.121 & 4.138 & 4.138  & 4.138 & 4.138  & 4.339   & 4.339     \\ \cline{2-11} 
                            & Ours with $\sigma_\tau$=4 & 10.124 & 5.487   & 4.040 & 3.672 & \textbf{3.364}  & 5.789 & 9.104  & 11.556  & 14.133    \\ \cline{2-11} 
                            & Ours with $\sigma_\tau$=8 & 10.124 & 5.775   & 4.141 & 3.729 & 3.571  & 6.199 & 11.1604 & 18.267 & 21.301     \\ \hline
\end{tabular}
\end{table*}
\subsection{Evaluation of the Collaborative Radio SLAM}
\label{sec:experiment_clam}
\begin{figure}

  \centering     

        \subfigure[Estimated trajectory for Test I]{

\label{fig:trajectory_test1}

        \includegraphics[width=0.49\textwidth]{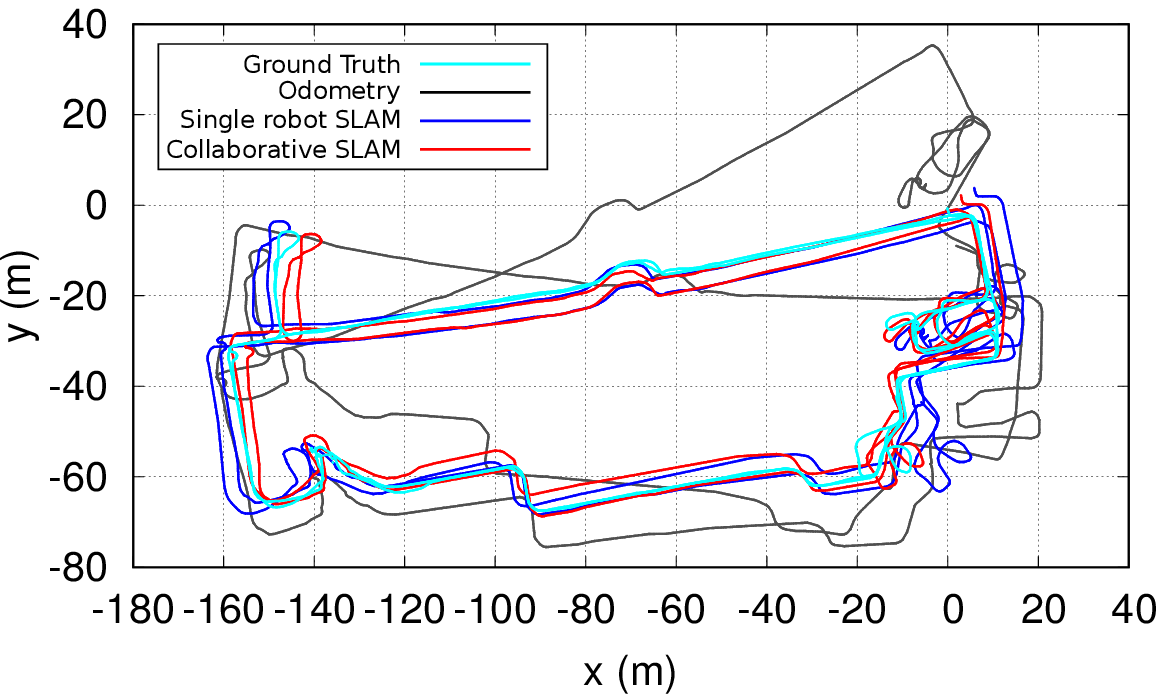}

        }

        \subfigure[Estimated trajectory for Test II]{

\label{fig:trajectory_test2}

        \includegraphics[width=0.49\textwidth]{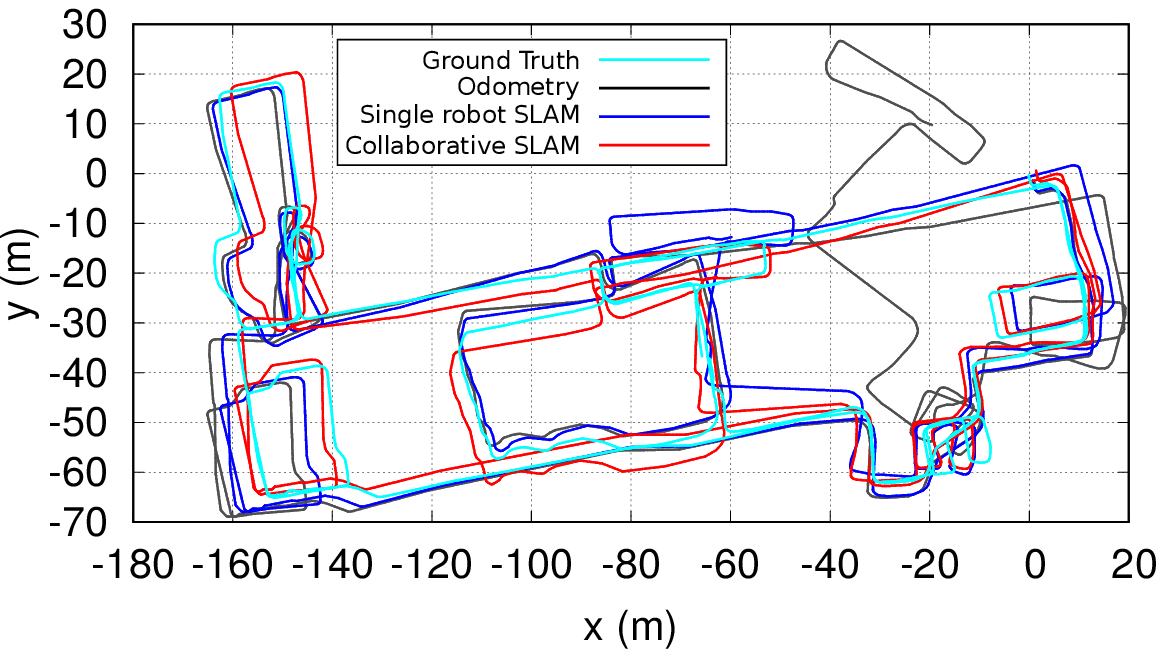}

        }
        \caption[Estimated trajectory]
{Estimated trajectory of different test sets under different approaches.}

\label{fig:accuracy_under_turning_detection}

\end{figure}

Next, we evaluated the performance of the proposed collaborative radio SLAM. 
Similar to the previous section, we set $\sigma_s=6$. We choose $\sigma_\tau=4$ for our proposed similarity measure.
The accuracy of optimized trajectory for Test I and Test II are summarized in Table \ref{table_slam_test1} and Table \ref{table_slam_test2}, respectively.
The results further verify that our proposed similarity measure outperforms cosine and Gaussian similarity measure.
In addition, the results show that the proposed solution helps to improve the accuracy of the trajectory estimation by the fusion of measurement from multiple robots.
For Test I, with the individual distance threshold $\nu_s=10$, our collaborative solution leads to a localization accuracy of 2.774m using $\nu_p=30$, 
which is an improvement of 14.9\%, when compared with a single robot SLAM (an accuracy of 3.261m). 
The same applies to Test II: we obtain a localization accuracy of 2.736m with $\nu_s=20$ and $\nu_p=10$, which is an improvement of 18.7\% when compared to the case without
collaborative SLAM (3.364m). 
Figure \ref{fig:trajectory_test1} and Figure \ref{fig:trajectory_test2} visualize the trajectory estimated with a single robot and our collaborative SLAM. 
The best choice of $\nu_p$ depends on the settings of $\nu_s$ (i.e., the parameter used to infer the individual constraints). 
In general, we should decrease $\nu_p$ with a large $\nu_s$ in order to get the best localization accuracy. 
A too large $\nu_s$ will lead to a reduction of the localization accuracy. 


\begin{table*}[]
\small
\centering
\caption{Trajectory estimation accuracy of Test I using collaborative SLAM.}
\label{table_slam_test1}
\begin{tabular}{|c|c|c|c|c|c|c|c|c|c|c|}
\hline
\multirow{2}{*}{$\nu_s$} & \multirow{2}{*}{Similarity measures} & \multicolumn{9}{c|}{Settings of parameter $\nu_p$}    \\ \cline{3-11} 
                          &                                      & 0 & 5 & 10 & 15 & 20 & 25 & 30 & 35 & 40 \\ \hline
\multirow{3}{*}{0}        & Cosine                               & 19.251  & 17.607  &  5.797  & 5.142   &  4.245 &  4.073  &  3.659  & 3.975   &  4.615  \\ \cline{2-11} 
                          & Gaussian                              & 19.251  &  6.385  &  4.626  & 3.590   &  3.356  & 3.087   & 3.974   & 8.624   &  18.928  \\ \cline{2-11} 
                          & Ours with $\sigma_\tau$=4           & 19.251  & 6.718  & 4.290    &  3.429  &  3.201  &  2.711  &  2.715  & 4.931   &  7.141  \\ \hline \hline
\multirow{3}{*}{5}        & Cosine                               & 4.876  &  5.183 &  4.199    &  4.196  &  4.089  &  3.974  & 3.773   &  3.924  & 4.625   \\ \cline{2-11} 
                          & Gaussian                              & 4.573 & 3.893  & 3.729      & 3.943   & 3.603   &  3.166  & 3.900   & 8.522   &  18.745  \\ \cline{2-11} 
                          & Ours with $\sigma_\tau$=4           &  4.503 & 4.004  & 3.450     & 3.426   &  3.209  &  2.771  &  2.787  &  4.646  &  6.804  \\ \hline \hline
\multirow{3}{*}{10}       & Cosine                               & 4.723  &  4.610 & 4.381     &  4.287  & 4.206   &  4.063  &  3.789  &  3.938  &  4.638  \\ \cline{2-11} 
                          & Gaussian                              & 4.375  & 4.013  & 3.724     & 3.702   & 3.445   & 3.208   &  3.943  &  8.346  & 18.683   \\ \cline{2-11} 
                          & Ours with $\sigma_\tau$=4           & 4.195  &  3.934 & 3.462     & 3.387   & 3.174   &  2.847  & \textbf{2.774}   & 4.576   &  6.716  \\ \hline \hline
\multirow{3}{*}{20}       & Cosine                               & 4.786  & 4.648  &  4.530    &  4.369  &  4.231  & 4.051   & 3.883   & 4.081   &  4.903  \\ \cline{2-11} 
                          & Gaussian                              &  4.062 & 3.513  &   3.302   &  3.254  &  3.096  &  3.286  &  4.600  & 8.713   & 18.491   \\ \cline{2-11} 
                          & Ours with $\sigma_\tau$=4           & 3.995  & 3.557  & 3.017     & 3.080   & 2.844   & 2.736   & 2.928   & 4.821   &  6.743  \\ \hline \hline
\multirow{3}{*}{30}       & Cosine                               &  4.782 & 4.729  & 4.192     &  3.939  &  3.918  &  3.892  &  4.046  &  4.724  &  5.591  \\ \cline{2-11} 
                          & Gaussian                              & 3.656  &3.215   &3.424      &3.902    &  4.271  & 4.867   &  6.272  &  9.882  & 18.818   \\ \cline{2-11} 
                          & Ours with $\sigma_\tau$=4           &  3.618 & 2.741  & 2.358     & 2.689   & 3.121   & 3.854   & 4.267   &  6.123  &  7.893  \\ \hline \hline
\multirow{3}{*}{40}       & Cosine                               & 4.563  & 4.189  & 3.979     &  4.208  &  4.359  & 4.687   & 5.083   & 5.976   &  6.598  \\ \cline{2-11} 
                          & Gaussian                              &  9.346 & 11.662  & 14.251  & 14.737   & 14.900  &  15.141  &  15.687  & 17.100   & 21.545   \\ \cline{2-11} 
                          & Ours with $\sigma_\tau$=4           & 3.858  & 4.667  &  5.345  &  5.828  &  6.190  & 6.763   &  7.128  & 8.673  &   9.971 \\ \hline
\end{tabular}
\end{table*}

\begin{table*}[]
\small
\centering
\caption{Trajectory estimation accuracy of Test II using our collaborative SLAM.}
\label{table_slam_test2}
\begin{tabular}{|c|c|c|c|c|c|c|c|c|c|c|}
\hline
\multirow{2}{*}{$\nu_s$} & \multirow{2}{*}{Similarity measures} & \multicolumn{9}{c|}{Settings of parameter $\nu_p$}    \\ \cline{3-11} 
                          &                                      & 0 & 5 & 10 & 15 & 20 & 25 & 30 & 35 & 40 \\ \hline
\multirow{3}{*}{0}        & Cosine                               & 10.124  &  9.224 & 5.426   & 5.297   & 5.07  & 4.124  &  4.381  & 4.608  & 5.465\\ \cline{2-11} 
                          & Gaussian                             & 10.124  & 7.690  &  4.591  &  4.182  & 4.044  &  4.341 & 6.262   & 11.536   &  22.994    \\ \cline{2-11} 
                          & Ours with $\sigma_\tau$=4           & 10.124  & 6.009  & 4.171   &  4.1322  & 3.910  & 4.366  & 4.914   &  7.746  & 10.166 \\ \hline \hline
\multirow{3}{*}{5}        & Cosine                               &  9.137 & 8.760  & 5.492   & 4.764   & 3.962  &4.032   & 4.297  &4.530   & 5.395 \\ \cline{2-11} 
                          & Gaussian                             &  5.916 & 5.384  &  4.138  & 3.355   &  3.289 &  3.707 & 5.454   &  10.757  &22.632 \\ \cline{2-11} 
                          & Ours with $\sigma_\tau$=4           & 5.487  & 4.729  & 4.407   & 3.882   & 3.377  & 3.682  &  4.114  &  6.835  & 9.541 \\ \hline \hline
\multirow{3}{*}{10}       & Cosine                               & 4.920  & 4.727  &  4.348  & 3.774   & 3.576  & 3.803  & 4.138   & 4.374  & 5.176\\ \cline{2-11} 
                          & Gaussian                             & 4.181  & 3.903  & 3.483   &  3.185  & 3.261  & 3.476  & 5.025   &  9.867  & 21.229 \\ \cline{2-11} 
                          & Ours with $\sigma_\tau$=4           & 4.040  &  3.959 &  3.183  &  3.259  &  3.256 & 3.512  & 3.905   & 6.289   & 8.640 \\ \hline \hline
\multirow{3}{*}{20}       & Cosine                              & 3.838  & 3.566  & 3.548  &  3.519  &  3.552  &4.170   & 4.490   &  4.859 & 5.651 \\ \cline{2-11} 
                          & Gaussian                             & 3.951  & 3.248  &  3.259  &  3.635  & 3.896  & 4.481  & 6.143   &  10.511  & 21.661 \\ \cline{2-11} 
                          & Ours with $\sigma_\tau$=4           & 3.364  &  3.3033 &  \textbf{2.736}  & 3.285   &  3.586 & 4.105  & 4.585   & 7.000   &9.267  \\ \hline \hline
\multirow{3}{*}{30}       & Cosine                               &  6.124 & 5.566  & 4.757   &  4.885  &4.985   & 5.065  &  5.365  &  5.861 &6.622 \\ \cline{2-11} 
                          & Gaussian                             &  11.97 & 9.564  &  8.178  & 7.944   & 8.254  & 8.855  & 10.246   & 13.630   &  22.577\\ \cline{2-11} 
                          & Ours with $\sigma_\tau$=4           & 9.104  & 7.547  & 6.592   & 6.146   & 6.411  & 6.962  &  7.459  & 9.464   &11.218  \\ \hline \hline
\multirow{3}{*}{40}       & Cosine                               & 12.016 &11.151   &   9.615 &  9.207  & 9.202  &  9.258&  9.360  & 9.542  & 10.065\\ \cline{2-11} 
                          & Gaussian                             & 24.913  & 22.916  & 20.832   & 20.599   &  20.721 & 20.927  & 21.365   &  22.480  &26.288 \\ \cline{2-11} 
                          & Ours with $\sigma_\tau$=4           & 14.133  & 12.423  & 10.852   &  10.481  & 10.609  & 11.064  & 11.355   & 12.626   & 13.910\\ \hline
\end{tabular}
\end{table*}

\section{Conclusion}
\label{sec:conclusion}
We presented a solution for collaborative simultaneous localization and mapping for multiple robots based on radio signals from the pervasive WiFi access points deployed in existing infrastructure. 
We applied a centralized implementation to fuse the odometry and WiFi radio signals collected from multiple robots. 
We further proposed a novel similarity measure that incorporates the received signal strength and detection likelihood of an access point. 
The performance of our approach is validated in a carpark with two sets of experiments. 
Our results show that our collaborative SLAM improves the trajectory estimation accuracy of a single robot SLAM by integrating the
 measurements from multiple robots.
We achieved an average positioning accuracy of less than 3 meters using our proposed collaborative SLAM and similarity model. 
In the future, we would like to investigate the possibility to fuse other sensory information, for example other sensor information (i.e., LiDARs), to improve the accuracy of trajectory estimation. 
Another direction is to develop a fully distributed solution to replace our centralized framework.
\bibliographystyle{IEEEtran}
\bibliography{bibSpace}

\end{document}